\pdfoutput=1

\documentclass[11pt]{article}

\usepackage[]{eacl2023}

\usepackage{times}
\usepackage{latexsym}

\usepackage[T1]{fontenc}

\usepackage[utf8]{inputenc}

\usepackage{xspace}
\usepackage{todonotes}
\usepackage{amsmath}
\usepackage{caption}
\usepackage{subcaption}
\usepackage{booktabs}
\usepackage{multirow}
\usepackage{tabularx}

\usepackage{tikz}
\usetikzlibrary{fadings}
\usetikzlibrary{mindmap,trees}
\usetikzlibrary{arrows,automata,shapes,positioning,shadows,trees}
\usetikzlibrary{shapes,snakes,shadows}
\usetikzlibrary{shapes.arrows}
\usetikzlibrary{calc,shapes, positioning}
\usetikzlibrary{decorations.text}
\usetikzlibrary{matrix,chains,positioning,decorations.pathreplacing,arrows}
\usetikzlibrary{bayesnet}
\usepackage[edges]{forest}
\usetikzlibrary{shadows.blur}
\usetikzlibrary{shapes.geometric}
\usepackage{pgfplots}
\usepackage{pgfplotstable}
\usepackage{pgfmath,pgffor}
\usepgfplotslibrary{colorbrewer}
\pgfplotsset{compat = 1.14, cycle list/Set1-8}
\usetikzlibrary{pgfplots.statistics, pgfplots.colorbrewer}
\pgfplotsset{compat=1.8}
\tikzstyle{edge}=[-latex',draw=black!90,shorten <=1pt,shorten >=1pt]
\tikzstyle{redge}=[latex'-,draw=black!90,shorten <=1pt,shorten >=1pt]
\tikzstyle{dedge}=[latex'-latex',draw=black!90,shorten <=1pt,shorten >=1pt]

\tikzstyle{block}=[draw, text width=5em,align=center,shape=rectangle, rounded corners, , align=center]
\tikzstyle{nobox}=[align=center]
\definecolor{emb}{RGB}{209,228,252}
\definecolor{hidden-blue}{RGB}{194,232,247}
\definecolor{hidden-orange}{RGB}{243,202,120}
\definecolor{hidden-yellow}{RGB}{242,244,193}
\definecolor{output-purple}{RGB}{219,203,231}
\definecolor{output-green}{RGB}{204,231,207}
\definecolor{hiddendraw}{RGB}{205, 44, 36}

\newcommand{\ie}{i.e.,\xspace}
\newcommand{\eg}{e.g.,\xspace}
\newcommand{\eat}[1]{}

\title{A Survey on Dynamic Neural Networks for Natural Language Processing}

\author{Canwen Xu, Julian McAuley \\
  University of California, San Diego \\
  \texttt{\{cxu,jmcauley\}@ucsd.edu}}

\begin{document}
\maketitle
\begin{abstract}
Effectively scaling
large 
Transformer models is a main 
driver of
recent 
advances
in natural language processing.
Dynamic neural networks, as an emerging research direction, are capable of scaling up neural networks with sub-linear increases in computation and time by dynamically adjusting their computational path based on the input.
Dynamic neural networks could be a promising solution to the growing parameter numbers of pretrained language models, allowing both model pretraining with trillions of parameters and faster inference on mobile devices.
In this survey, we 
summarize the progress of three types of dynamic neural networks in NLP:
skimming, mixture of experts, and early exit.  
We also highlight current challenges in dynamic neural networks and directions for future research.

\end{abstract}

\section{Introduction}
Scaling up 
model capacity is an obvious yet effective approach for better performance in natural language processing (NLP) tasks~\citep{gpt3,kaplan2020scaling,ghorbani2021scaling,zhou2020evaluating}. However, 
the 
resulting
increase in computational complexity and memory consumption becomes a bottleneck for
scaling, making these models hard to train and use. On the other hand, it 
is not necessary
to allocate the same amount of computation to all instances. For example, categorizing ``I love you'' as a positive sentence 
does not require a model containing dozens of Transformer layers. To resolve the aforementioned problems, \textit{dynamic neural networks} have 
been a significant thrust of recent research in NLP. Dynamic networks can adjust their computational path based on the input for better efficiency, making it possible to train models with trillions of parameters and accelerate models in a low-resource setting.

\begin{figure}[t]
\centering
  \includegraphics[width=1\columnwidth]{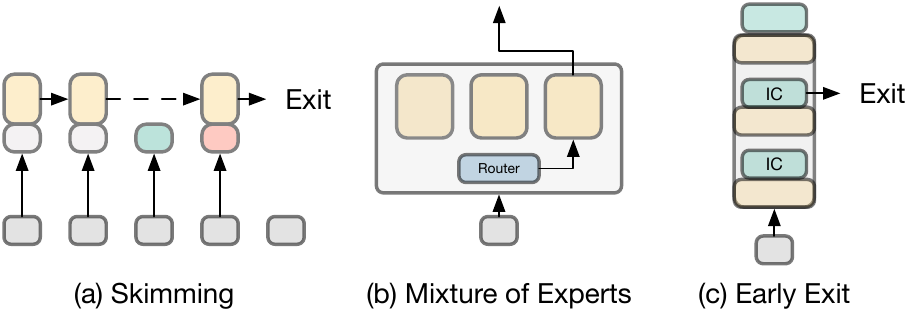}
  \caption{The three types of dynamic neural networks summarized in this paper. They dynamically adjust computation \textit{timewise}, \textit{widthwise} and \textit{depthwise}, respectively.}
  \label{fig:dynamic}
\end{figure}

In this survey, we review the latest state of research on three types of dynamic neural networks that have been adopted in NLP: \textit{skimming}, \textit{mixtures of experts} (MoE), and \textit{early exit}, as illustrated in Figure~\ref{fig:dynamic}. These three types of techniques share a common idea of dynamically adjusting computation with respect to input, to save computation through bypassing unnecessary modules in a large neural network. However, they implement the goal via different approaches. \textbf{Skimming} was well-researched in the era of recurrent neural networks (RNN). Skimming models save computation \textit{timewise} by dynamically allocating computation to different time steps, based on the %
input tokens.
Since RNN models process the input sequence recurrently, it allows skimming models to achieve a substantial acceleration, especially when the sequence is long~\citep{li2019teach}. Different from RNN, recent works on Transformers skip tokens between layers instead of time steps.

For Transformer models~\citep{transformer,bert,albert,gpt3}, the input tokens are fed into the model in parallel, while models have dozens of Transformer layers. This motivates the development of MoE and early exit.
\textbf{MoE} horizontally extends a feedforward neural network (FFNN) with multiple sub-networks. During inference, only one or a few of these sub-networks will be activated for computation, thus can save \textit{widthwise} computation. \textbf{Early exit}, on the other hand, terminates inference at an early layer, without exhausting full computational capacity, thus saves \textit{depthwise} computation. Early exit techniques often insert a series of lightweight classifiers which help decide when to exit, based on an exit strategy. 

Note that this stream of works is distinct from static model acceleration, which is often referred to as \textit{model compression}, including knowledge distillation, weight sharing, pruning and quantization~\citep{sanh2019distilbert,bot,albert,zafrir2019q8bert,xu2021beyond} (etc., see another survey~\citep{xu2022survey}). The major difference is that the computational path in a statically compressed model does not condition on the input and is invariable for all examples in inference. These two streams of research are in fact orthogonal and recent works \citet{schwartz2020right}, \citet{fastbert} and \citet{leebert} have shown that static and dynamic approaches can be combined for even faster inference and better performance.

To summarize, our contribution is two-fold: (1) We review the latest studies on the topic of dynamic neural networks for NLP by providing a comprehensive comparison and organize them with a new taxonomy, as shown in Figure~\ref{fig:dynamic}.
(2) We analyze current challenges in dynamic neural networks and point out directions for future research.

\begin{figure}[t]
  \centering
\begin{forest}
  forked edges,
  for tree={
    grow=east,
    reversed=true, %
    anchor=base west,
    parent anchor=east,
    child anchor=west,
    base=left,
    font=\footnotesize,
    rectangle,
    draw=hiddendraw,
    rounded corners, 
    align=left,
    minimum width=2.5em,
    minimum height=1.2em,
    s sep=6pt,
    inner xsep=3pt,
    inner ysep=1pt,
  },
  where level=1{font=\scriptsize}{},
  where level=2{font=\scriptsize}{},
  where level=3{font=\scriptsize}{},
  where level=4{font=\scriptsize}{},
  where level=5{font=\scriptsize}{},
  [Dynamic \\ Models \\ for NLP
    [Skimming
      [Skipping and Early Stopping]
      [Computation Reduction]
      [Dynamic Hierarchical RNN]
    ]
    [Mixture of Experts
      [Learned Routing]
      [Unlearnable Routing]
    ]
    [Early Exit
      [Confidence-based]
      [Ensemble-based]
      [Learning-based]
      [Cascading]
    ]
  ]
\end{forest}
\caption{Taxonomy of dynamic neural networks for NLP.}
\label{fig:tax}
\end{figure}
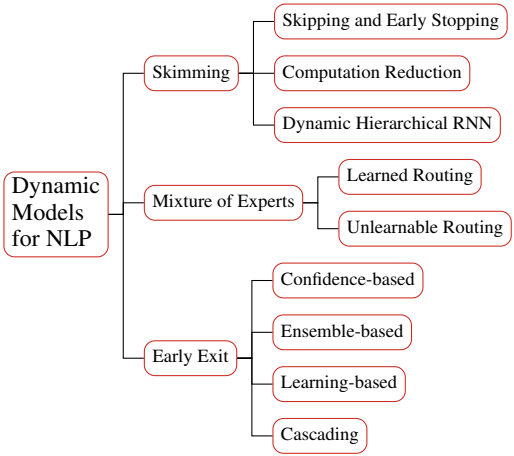

\begin{table*}[t]
\centering
\resizebox{1.\linewidth}{!}{
\begin{tabular}{lll}
\toprule
\textbf{Method}  & \textbf{Decision based on} & \textbf{Operation options} \\
\midrule
LSTM-Jump~\citep{yu2017learning}     & hidden states  & skip multiple steps; stop reading \\
Skip RNN~\citep{campos2018skip} & states of the update gate; hidden states & skip a single step \\
ReasoNet~\citep{shen2017reasonet} & hidden states & stop reading \\
Jumper~\citep{liu2018jumper} & input sentence; hidden states & stop reading \\
RIM~\citep{li2019teach} & input sentence; hidden states & skip a single step; stop reading\\
\citet{yu2018fast} & hidden states & skip multiple steps; stop reading; re-read \\
LSTM-Shuttle~\citep{fu2018speed} & hidden states & skip multiple steps; jump back multiples steps \\
Struc. Jump-LSTM~\citep{hansen2019neural} & hidden states & stop reading; jump to next (,;) or (.!?) \\
\midrule
PoWER~\cite{powerbert} & attention & drop tokens\\
TR-BERT~\cite{trbert} & hidden states & forward tokens \\
LAT~\cite{lat} & attention & forward tokens \\
 LTP~\cite{ltp} & attention & drop tokens \\
 Transkimmer~\cite{transkimmer} & hidden states & forward tokens \\

\midrule
VCRNN~\citep{jernite2017variable} & input token; hidden states & partial update with zero-masked weights\\
Skim-RNN~\citep{seo2018neural} & input token; hidden states & partial update with a small RNN \\
\midrule
HM-RNN~\citep{chung2017hierarchical} & states of the gates & skip a single step; ``flush'' \\
FHRNN~\citep{ke2018focused} & query; hidden states & update the upper RNN layer \\

\bottomrule
\end{tabular}}
\caption{A summary of skimming techniques.}
\label{tab:skim}
\end{table*}

\section{Skimming}

Skimming techniques, as summarized in Table~\ref{tab:skim},
skip some time steps or allocate different computation on them. Intuitively, skimming matches how human beings efficiently read text and extract information from it~\citep{li2019teach}. By emphasizing the important information within a sequence and ignoring parts with little importance, skimming helps the model achieve faster inference speed and better capture long-term dependencies. The three categories of skimming are \textit{skipping and early stopping}, \textit{computation reduction}, and \textit{dynamic hierarchical RNN}, corresponding with three motivations: to skip unimportant input, to allocate less computation to unimportant input, and to increase computation to important input only.

\paragraph{Skipping and Early Stopping} Skipping and early stopping aim to improve efficiency for a long sequence by skipping some tokens or stopping reading early. LSTM-Jump~\citep{yu2017learning} is a skipping mechanism to ignore irrelevant information for natural language understanding (NLU). At each step, the current states are used to compute a ``jumping softmax'', which decides how many steps to jump forward and whether to stop reading. LSTM-Jump employs policy gradient to train the model to make non-differentiable discrete jumping decisions. The reward is a binary function which rewards a correct prediction and penalizes an incorrect prediction of the label. Compared to a standard LSTM, LSTM-Jump achieves better accuracy with up to 6$\times$ speed-up. Skip RNN~\citep{campos2018skip} introduces a binary gate to learn whether to skip a state update. If the gate decides to skip a time step, the hidden states will be directly copied without any update.

To stop reading early as needed, ReasoNet~\citep{shen2017reasonet} introduces a terminal state which decides whether to terminate early for machine reading comprehension on each time step at the token level. Jumper~\citep{liu2018jumper} first splits a paragraph to several sub-sentences and encodes them into sentence embeddings. They then apply early stopping at a sentence when the policy network decides to stop reading. 
\citet{li2019teach} use eye-tracking devices and confirm that skipping and early stopping are common when humans read text. They propose Reading Inspired Model to mimic the behaviors of humans, which allows the model to decide whether to skip a single time step or stop reading early.
\citet{yu2018fast} add a re-reading operation to LSTM-Jump~\citep{yu2017learning} which allows the model to stay on the current time step, allocating more computation to important information.

The aforementioned techniques can only go forward, which makes it impossible to regret if having jumped over important information. LSTM-Shuttle~\citep{fu2018speed} proposes a bidirectional shuttling mechanism, which can jump multiple time steps both forward and backward, allowing the model to ignore unimportant information and recover lost information if needed.

Structural information that naturally exists in sentences can also play a role in skimming. Structural Jump-LSTM~\citep{hansen2019neural} can jump to the next word, next sub-sentence separator (a comma or colon), next sentence end symbols (a period, exclamation mark or question mark), or to the end of the text (\ie stop reading).

In the era of Transformers, there have been works attempting to reduce computation by either skip tokens at higher layers or forward tokens to higher layers. The PoWER-BERT model~\citep{powerbert} reduces the number of tokens processed by each Transformer layer based on their attention scores. The number of tokens to be dropped, referred to as the "schedule," is optimized by combining the sparsity of a soft mask layer with the original loss function. This results in an improved balance between accuracy and processing time. TR-BERT~\citep{trbert} uses a dynamic approach to determine which tokens to skip, using reinforcement learning to train the model with a reward system that prioritizes classifier confidence while also penalizing the number of tokens retained. In contrast to PoWER-BERT, TR-BERT passes the skipped tokens to the final layer rather than discarding them. The Length-Adaptive Transformer (LAT, \citealp{lat}) utilizes LengthDrop to randomly skip tokens during pretraining, aiming to close the gap between pretraining and fine-tuning. The schedule for LAT is found through an evolutionary search algorithm. LTP~\citep{ltp} trains a threshold for each Transformer layer, instead of following a predetermined schedule. It simply drops tokens with attention scores lower than the learned threshold. Transkimmer~\citep{transkimmer} incorporates a skim predictor module, consisting of a small MLP and Gumbel-Softmax reparameterization, before each layer. This module outputs a mask to determine whether a token should be dropped, and a skim loss is used to optimize the ratio of skipped tokens to total tokens, promoting sparsity.

\paragraph{Computation Reduction} Different from skipping, computation reduction applies a reduced computational workload for some time steps instead of skipping it completely. VCRNN~\citep{jernite2017variable} explores a scheduler to decide which proportion of computation to use for each time step. Upon making the decision, only the corresponding proportion of the weight matrix will be used to update the hidden states while the rest part of the weight matrix will be masked out with zero.

Instead of using part of weights to update the hidden states, Skim-RNN~\citep{seo2018neural} has a big RNN and a separate small RNN. At each time step, the model decides whether to read or skim based on hidden states from the last time step and the input token. If the model decides to skim, the small RNN will update only a fraction of the hidden states. Otherwise, a regular full update will be conducted by the big RNN.

\paragraph{Dynamic Hierarchical RNN} Different from the aforementioned two categories of skimming, dynamic hierarchical RNN can \textit{increase} computation by calling the upper layer RNN when needed.
HM-RNN~\citep{chung2017hierarchical} automatically discovers the hierarchical multi-scale structure in the data for a hierarchical RNN architecture. In addition to the update and copy operations as in Skip RNN~\citep{campos2018skip}, they add a flush operation which ejects the summarized representation of the current time step to the upper layer and re-initializes the states for the next time step.

In question answering, only a small portion of tokens are relevant and can be used to answer the question while the rest can be safely skimmed.
Based on this observation, Focused Hierarchical RNN~\citep{ke2018focused} aims to only pick up information that is relevant to the query for question answering. It applies a binary gate to control whether to update the upper layer of the RNN, based on the current hidden states of the lower-level RNN and the question embedding.



\section{Mixture of Experts}

\begin{table*}[t]
\centering
\resizebox{1.\linewidth}{!}{
\begin{tabular}{llcl}
\toprule
\textbf{Method} & \textbf{Base Model} & \textbf{Sparsity} & \textbf{Load Balance} \\
\midrule
Sparsely~\citep{shazeer2017outrageously} & LSTM & top-$k$ & auxiliary loss \\
GShard~\citep{gshard} & Transformer (NMT) & top-2 & expert capacity; local group dispatching; auxiliary loss; random routing \\
Switch~\citep{fedus2021switch} & Transformer (T5) & top-1 & expert capacity; auxiliary loss \\
BASE~\citep{lewis2021base} & Transformer (GPT) & top-1 & linear assignment \\
M6-T~\citep{yang2021m6} & Transformer (M6) & $k$ top-1 & expert capacity \\
DTS~\citep{nie2021dense} & Transformer (GPT) & dynamic & sparsity scheduler \\
\midrule
Hash~\citep{roller2021hash} & Transformer & hash & deterministic hash \\
THOR~\citep{zuo2022taming} & Transformer (NMT) & random & random selection \\
\bottomrule
\end{tabular}
}
\caption{A summary of Mixture of Experts (MoE) methods.}
\label{tab:moe}
\end{table*}

Increasing the number of parameters in a model often leads to increased computation and memory consumption.
To take the advantages of parameter scaling without a proportional increase in computation, mixture of experts (MoE)~\citep{jacobs1991adaptive} is introduced to large language models, as summarized in Table~\ref{tab:moe}. In these models, a layer typically contains multiple sub-networks (\ie ``experts''). During inference, only part of these experts will be activated on a per-example basis.

The key element of MoE methods is the routing mechanism.
The routing mechanism has to be lightweight, not to significantly slower the speed of the model. We categorize MoE methods into two groups: \textit{learned routing} and \textit{unlearnable routing}.  Learned routing often requires some load balancing mechanisms to ensure that all experts are trained with enough examples thus are useful during inference. Unlearnable routing usually slightly underperforms learned routing but does not require complicated load balancing.

\paragraph{MoE Layers with Learned Routing} A straightforward idea to implement MoE is to learn a router to allocate inputs to experts. Sparsely-Gated MoE layer~\citep{shazeer2017outrageously} contains up to thousands of feed-forward sub-networks with a trainable gating network which determines a sparse combination of these experts to use for each example. There are two major challenges to address: \textbf{(1) Sparsity.} The gating network predicts a softmax weight for the experts based on the input.
The gating network is trained by simple back-propagation, together with other parts of the model. Then, only the top-$k$ experts in the layer will be activated based on the softmax prediction of the gating network. They insert one MoE layer between stacked LSTM layers and achieve improvement on language modeling and machine translation tasks.
\textbf{(2) Load balancing.} \citet{shazeer2017outrageously} observe a self-reinforcing phenomenon that the gating network tends to converge to a state where it always produces large weights for the same few experts. They resolve the problem by defining the importance of an expert relative to a batch of training examples to be batch-wise sum of the \textit{gate values} for that expert. Then, they introduce an additional loss, the square of the coefficient of variation of the set of importance values, to encourage a more balanced update during training. Besides encouraging a balanced update, the authors also introduce a loss function with a smooth estimator that estimate the number of examples assigned to each expert for a batch of inputs, to encourage experts to receive roughly equal \textit{numbers of training examples}.

GShard~\citep{gshard} enables scaling up multilingual neural machine translation Transformer beyond 600 billion parameters. It adapts Sparsely-Gated MoE~\citep{shazeer2017outrageously} to Transformer~\citep{transformer} by replacing every other feed forward layer with an MoE layer, which routes to top-2 experts. When scaling to multiple devices, the MoE layer is sharded across devices, \ie each device has different allocated experts, while all other layers are replicated. To achieve workload balance, GShard employs a threshold, namely expert capacity, to limit the maximum number of tokens processed by one single expert. They also introduce a local group dispatching mechanism, which partitions all tokens in a training batch evenly into groups to be processed independently in parallel, to balance the overall workload. Following \citep{shazeer2017outrageously}, they use an additional loss to enforce even allocation for experts. Additionally, they propose a random routing mechanism, which only routes to the second-best expert with  probability proportional to its weight, to simplify sparse training.

Switch Transformer~\citep{fedus2021switch} aims to simplify the Sparsely-Gated MoE~\citep{shazeer2017outrageously} for efficiency and performance. They propose a Switch Layer which only routes to one expert at a time, to reduce gating computation, batch size and communication costs. Switch Transformer inherits expert capacity and an auxiliary load balancing loss from GShard~\citep{gshard}. Combined with low-precision training, compared to T5-Base and T5-Large~\citep{t5}, Switch Transformer obtains up to 7$\times$ increases in pretraining speed with the same computational resources. They further scale Switch Transformer to more than 1.5 trillion parameters and achieve 4$\times$ speed-up over T5-XXL. 

The Balanced Assignment of Sparse Experts (BASE) layer~\citep{lewis2021base} formulates token-to-expert allocation as a linear assignment problem and solves it with the auction algorithm~\citep{bertsekas1992auction}. This allows an optimal assignment in which each expert receives an equal number of tokens, improving efficiency and getting rid of the expert capacity and auxiliary loss in previous works. The experiments show that BASE layers are more efficient for training compared to Sparsely-Gated MoE layers~\citep{shazeer2017outrageously} and Switch Layers~\citep{fedus2021switch}, and can successfully learn a good balanced routing without any auxiliary balancing loss.

M6~\citep{lin2021m6} is a multi-modal multitask Transformer, trained in the same way as Switch Transformer~\citep{fedus2021switch}, scaling up to 100B parameters. Following this, M6-T~\citep{yang2021m6} splits experts into $k$ prototypes (\ie groups of experts). In each forward pass, each token is sent to the $k$ prototypes, within which the top-1 routing is done locally. The experiments demonstrate this ``$k$ top-1'' strategy outperforms the top-1 routing in Switch Transformer~\citep{fedus2021switch} while being more computation-efficient than ``top-$k$'' routing. They also claim that the load balancing loss may be ineffective for improving the performance of an MoE model, although it can indeed help balance the workload. They subsequently train a 1 trillion parameter model with the finding.

Dense-to-Sparse gate~\citep{nie2021dense} begins as a dense gate that routes tokens to all experts then gradually learns to become sparser and route tokens to fewer experts, demonstrating higher training efficiency in experiments. 
Their experiments confirm the finding in \citet{yang2021m6} that an auxiliary load balancing loss does not improve the model performance.

\paragraph{MoE Layer with Unlearnable Routing} Although learning-based routing has shown effectiveness only with the help of complicated load balancing mechanisms, recent studies have attempted to get rid of those.
Hash Layer~\citep{roller2021hash} simplifies routing by using a parameter-free hashing function to route tokens to specific experts. This design eliminates the need for a load balancing loss and sophisticated assignment algorithms. They also study the performance of different hashing techniques, hash sizes and input features, and conclude that balanced and random hashes focused
on the most local features work best. The experiments show that a Hash Layer achieves comparable performance with a Switch Layer~\citep{fedus2021switch} and BASE Layer~\citep{lewis2021base}.

THOR~\citep{zuo2022taming} is a special form of MoE layer, which completely discards the conditional routing mechanism and instead optimizes the consistency between a randomly selected pair of experts. During inference, one expert will be randomly selected to be activated.

\paragraph{Applications and Analysis}  GLaM~\citep{du2021glam} trains a family of GPT-style language models with up to 1.2 trillion parameters using GShard~\citep{gshard}.
CPM-2~\citep{zhang2022cpm} trains a large Chinese language model with 198 billion parameters with BASE layers~\citep{lewis2021base}. 

\citet{artetxe2021efficient} conduct a detailed empirical study of how autoregressive MoE language models scale compared to dense models. They find MoEs to be substantially more efficient with the exception of fine-tuning. MoE models can match the performance of dense models with 25\% of computation in a low-resource setting. Although the advantage fades at scale, their largest MoE model with 1.1 trillion parameters can consistently outperform its dense counterpart with the same amount of computation. \citet{clark2022unified} examine the scaling law of BASE Layer~\citep{lewis2021base}, Hash Layer~\citep{roller2021hash} and earlier Reinforcement Learning-based routing algorithms providing suggestions for best-practices in training MoE models.

\citet{zhang2021moefication} propose MoEfication to split feedforward neural networks (FFNN) in a trained large model to experts. They find that a T5-Large~\citep{t5} model with 700 million parameters only activates 5\% neurons for 80\% inputs on a downstream task, indicating high redundancy within large pretrained language models. To transform a pretrained language model to an MoE model, they first construct a co-activation graph for each FFNN and then divide the graph into subgraphs with strong internal connections with graph partitioning algorithm. Each subgraph forms an expert. They train a router with oracle best routing for training data. Then, they further fine-tune the resulted model for better performance.

\begin{table*}[t]
\centering
\resizebox{1.\linewidth}{!}{
\begin{tabular}{lll}
\toprule
\textbf{Method} & \textbf{Internal classifier training} & \textbf{Exit criterion} \\
\midrule
DeeBERT~\citep{deebert} & two-stage; sum of CE loss & entropy $<\theta$ \\
RightTool~\citep{schwartz2020right} & joint; sum of CE loss & calibrated max class probability $>\theta$ \\
FastBERT~\citep{fastbert} & two-stage; self-distillation & entropy $<\theta$   \\
RomeBERT~\citep{geng2021romebert} & joint; self-distillation + GR & entropy $<\theta$   \\
SkipBERT~\shortcite{skipbert} & joint; weighted sum of CE + KD & max class probability $>\theta$ \\

\midrule
PABEE~\citep{pabee} & joint; weighted sum of CE loss & patience (\#consistent prediction $>\theta$ ) \\
Voting~\citep{sun2021early} & joint; sum of CE + diversity loss & accumulated votes $>\theta$  \\
LeeBERT~\citep{leebert} & joint; auto-weighted sum of CE + KD loss & patience (\#consistent prediction $>\theta$ ) \\
Past-Future~\citep{liao2021global} & joint; weighted sum of CE + imitation learning & entropy $<\theta$  \\
PCEE-BERT~\shortcite{pceebert} & joint; weighted sum of CE & patience (\#consistent IC confidence $>\theta$) \\
\midrule
BERxiT~\citep{xin2021berxit} & alternate; sum of CE loss & estimated confidence $>\theta$  \\
CAT~\citep{schuster2021consistent} & joint; avg. of CE loss & estimated conformity $>\theta$  \\
\midrule
CascadeBERT~\citep{cascadebert} & standard model FT with confidence calibration & calibrated max class probability $>\theta$  \\
\bottomrule
\end{tabular}
}
\caption{A summary of early exit methods. $\theta$ is a predefined threshold for exiting. This table is extended from a table in \citet{xu2022survey}.}
\label{tab:ee}
\end{table*}

\section{Early Exit}
\label{sec:ee}
Early exit techniques aim to terminate model inference in early layers, to save computation and sometimes improve performance by resolving the overthinking problem~\citep{shallowdeep}, \ie possible performance degradation at a later layer. It can be useful especially in the era of pretrained language models (PLM), since increasing the size of PLMs can often lead to better performance, although a smaller model can already predict most examples (\ie ``easy examples'') correctly.

The main idea of early exit is to exit inference at an earlier layer, rather than the last layer. Early exit often involves a series of internal classifiers inserted into a large network, providing signals for early exiting. The core of early exit methods is the exit criterion. Based on their exit strategies, we categorize the early exit methods into three classes: confidence-based, ensemble-based and learning-based, as listed in Table~\ref{tab:ee}.

Despite better performance, speed and adversarial robustness~\citep{pabee}, an additional benefit is that the speed-accuracy trade-off can be adjusted as needed by tuning the exit threshold ($\theta$ in Table~\ref{tab:ee}), without the need of retraining the model. 
A main drawback is that early exit is often applied on a per-instance basis, meaning that to maximize the speed-up ratio, a small batch size (often 1) has to be used.

\paragraph{Confidence-based Early Exit} Early works for early exit in computer vision~\citep{park2015big,branchynet,shallowdeep} often fall into this category. They define a metric as the proxy for confidence of a model prediction. The model exits early when the confidence hits a predefined threshold. 
DeeBERT~\citep{deebert} applies BranchyNet~\citep{branchynet} to BERT inference. The training for DeeBERT is two-stage: they first train BERT on downstream tasks following standard fine-tuning. Then, they freeze the parameters of the Transformer and insert a linear classifier (\ie internal classifier) after each Transformer layer. They train the classifiers by minimizing the sum of their cross-entropy loss. For inference, the model exits early when an internal classifier outputs a prediction probability distribution that has an entropy lower than a predefined threshold. RightTool~\citep{schwartz2020right} jointly fine-tunes BERT with internal classifiers. They use the temperature-calibrated maximum class probability as confidence. FastBERT~\citep{fastbert} first trains the BERT backbone and the final classifier. Then, they distill the final classifier layer to the internal classifiers~\citep{hinton2015distilling}. For inference, the model exits when the entropy of a prediction is below the threshold. RomeBERT~\citep{geng2021romebert} provides a simple fix for learning internal classifiers efficiently. Besides self-distillation as in FastBERT, they propose gradient regularization (GR) to facilitate  distillation. SkipBERT~\citep{skipbert} caches pre-computed representation of text chunks to replace lower BERT layers and uses confidence-based early exit for higher layers to achieve maximum acceleration.

\paragraph{Ensemble-based Early Exit} One drawback in confidence-based early exit is wasted computation. That is to say, if the confidence of an internal classifier does not satisfy the exit criterion, it will be disregarded. Ensemble-based early exit recycles these predictions and considers output from multiple internal classifiers to make better predictions. Based on the similarity between overfitting and overthinking, PABEE~\citep{pabee} borrows early stopping from model training. They first jointly train the internal classifiers with BERT by a weighted sum of cross-entropy losses that assigns larger weights for upper classifiers. For inference, the model exits when $k$ consecutive internal classifiers make the same prediction. 
Other than
improvement on performance and efficiency, they  find that PABEE can improve adversarial robustness, which they attribute to the ensemble effect. \citet{sun2021early} further introduce a diversity loss that encourages internal classifiers to have a diverse predicted probability distribution. They propose a voting mechanism to ensemble the internal classifiers by exiting early when a class has accumulated more votes than the threshold. 
Interestingly, LeeBERT~\citep{leebert} adopts the opposite strategy: they promote consistency across  internal classifiers by distilling them to each other. However, they introduce a learnable weight for the cross-entropy loss of each classifier and the distillation loss between each pair. They optimize these weights by a cross-level optimization algorithm. They adopt PABEE's patience-based strategy for exiting.
\citet{liao2021global} train linear transformation layers called ``imitation learners'', to approximate the hidden states of future layers based on current hidden states. For inference, the prediction after each layer is calculated by mixing the past predictions and the future predictions of the imitation learners. Entropy is used as the exit criterion. PCEE-BERT~\citep{pceebert} borrows from both ensemble-based exit and confidence-based methods. The inference is terminated when multiple layers are confident.

\paragraph{Learning-based Early Exit} Another stream of research is to \textit{learn} a criterion for early exiting.
BERxiT~\citep{xin2021berxit} alternates between joint fine-tuning and two-stage fine-tuning by freezing parameters of Transformer and the final classifier for even-numbered iterations and unfreezing them for odd-numbered iterations. They also train a linear layer called a \textit{learning-to-exit} (LTE) module to predict whether the current internal classifier makes the correct prediction. It takes the hidden states as input and outputs a confidence score, which is used to decide whether to exit. CAT~\citep{schuster2021consistent} introduces a ``meta consistency classifier'' to predict whether the output of an internal classifier conforms to the final classifier and exits when the consistency classifier predicts a certain level of conformity.

\paragraph{Cascading} Cascading can be seen as a special form of early exit, performed at the model level. \citet{cascadebert} find that shallow features and internal classifiers in the first few layers of BERT utilized by early exit methods like DeeBERT~\citep{deebert} are not sufficient and reliable, underperforming a fine-tuned BERT with the same number of layers. Therefore, they propose to use a suite of complete models with different numbers of layers for cascading. CascadeBERT executes models one by one, from the smallest to the largest. It stops when a model outputs a confidence score (calibrated maximum class probability) that reaches the threshold.

\paragraph{Applications} Although early exit is originally developed for classification, there have been works extending it to more tasks and settings. \citet{li2021accelerating} propose Token-Level Early-Exit that targets early exiting for sequence labeling. They use the maximum class probability as confidence on a per-token basis. Once the confidence hits the threshold, the hidden states of the corresponding tokens will be frozen and directly copied to upper layers. These exited tokens will not attend to other tokens at upper layers but can still be attended by other tokens. The model completely exits when every token exits. A similar idea is also presented in \citet{elbayad2020depth} and \citet{liu2021faster} where hidden states of some positions can be frozen and directly copied to upper layers, although the former is focused on generation and the latter is for classification. \citet{xin2020early} apply DeeBERT~\citep{deebert} to document ranking and set different thresholds to the negative and positive classes for early exiting, to accommodate the imbalanced class distribution in document ranking. ELUE~\citep{liu2021towards} is a benchmark which evaluates the Pareto Front of early exit models on the FLOPs-performance plane. They provide a BERT-like baseline with jointly pretrained internal classifiers, to mitigate the gap between pretraining and fine-tuning.

\section{Challenges and Future Directions}

\paragraph{Evaluation} Evaluating dynamic neural networks can be difficult since we cannot pre-define a few break points to compare different methods at the exact same amount of computation or time. ELUE score~\citep{liu2021towards} may be a promising solution to this problem by considering both computation and performance, depicting the Pareto Front.
Besides, different works have different calculation for speed-up ratio. For example, some works use the ratio of layers involved in computation to estimate speed-up ratio~\citep{pabee,sun2021early,liao2021global}. This can be misleading since internal classifiers introduce extra computational costs, especially when more complicated mechanism introduced, \eg future-layer imitation~\citep{liao2021global}. Also, the reported speed of MoE models, greatly differs on different hardware and distribution settings, making it hard to compare across papers.

\paragraph{Data Parallelism} One drawback of dynamic neural networks is their inefficiency on data parallelism. To be specific, MoE methods introduce extra communication costs for dynamic routing and could be a bottleneck for efficiency. Skimming and early exit methods often employ an ``online inference'' setting where the batch size is fixed to 1, to achieve maximum acceleration. However, for batched inference, the efficiency of these methods will drastically degrade, since the already-exited instances will have to wait all instances to exit, which causes a low parallelism and low utilization of GPU. 

\paragraph{Optimized Runtime} Since dynamic neural networks are an emerging type of neural networks, most hardware and libraries are not well-optimized for these models. For example, sparse matrix multiplication in MoE needs specialized hardware and software support to achieve its theoretical efficiency. Also, current dynamic neural networks are often implemented in eager execution, which prevents them from low-level optimization of graph execution. There have been works exploring optimized runtime for MoE~\citep{meshtf,jia2020whale,he2021fastmoe,rajbhandari2022deepspeed} and early exit~\citep{paul2019hardware} while more to be done in the future.

\paragraph{Theoretical Analysis and Support} While the dynamic neural networks have demonstrated empirical improvement over static counterparts, dynamic networks are not solidly backed by theoretical analysis. For example, the theoretical analysis in PABEE~\citep{pabee} is based on an assumption that internal classifiers are independent to each other, which is unrealistic. More research should be done from the perspective of optimization and effect of data distribution on dynamic neural networks.

\paragraph{Explainability} The decision-making process of the dynamic neural networks can be important to explain the model prediction and even understand more fundamental research questions in machine learning, including scaling law and generalization. Can we use skimming to explain sequence classification? Is it consistent with attention-based explanation~\citep{xu2015show}?  What does each expert in MoE learn and what makes them different? Why does a lower internal classifier make different prediction from an upper classifier despite equally trained with the same objective? These questions warrant further exploration, from both data and model perspectives.

\section*{Limitations}
A limitation of this survey is that we do not draw a direct quantitative comparison for the methods surveyed in this paper since different methods have their own accuracy-speed curves, with their own unique limitations (e.g., many early exit methods can only handle a batch size of 1). Also, we do not discuss some works in depth and in detail due to space limit.

\bibliographystyle{acl_natbib}
\bibliography{anthology,custom}

\end{document}